\documentclass{article}
\usepackage{spconf,amsmath,graphicx}
\usepackage{amssymb}
\usepackage{enumitem}
\usepackage{multirow}
\usepackage[table]{xcolor} 
\usepackage{url}
\usepackage{makecell}
\usepackage{comment}
\usepackage{highlight}
\usepackage{fancybox}
\usepackage{tcolorbox}
\usepackage{float}
\usepackage{stfloats}
\usepackage[font=footnotesize]{caption}
\newtcolorbox{mytcolorbox}{fontupper=\footnotesize,boxrule=1pt,boxsep=0pt,left=4pt,right=4pt,top=4pt,bottom=4pt}
\usepackage{fancyhdr}
\renewcommand{\opacity}{50}

\usepackage[belowskip=-10pt,aboveskip=2pt]{caption}
\newcolumntype{Y}{>{\centering\arraybackslash}X}
\renewcommand{\arraystretch}{1.5}

\ninept
\newlength\myindent
\def\Width{0\kern2\tabcolsep\ldots\kern1\tabcolsep0}
\usepackage{etoolbox}
\newcommand{\zerodisplayskips}{%
  \setlength{\abovedisplayskip}{3pt}
  \setlength{\belowdisplayskip}{3pt}
  \setlength{\abovedisplayshortskip}{3pt}
  \setlength{\belowdisplayshortskip}{3pt}}
\appto{\normalsize}{\zerodisplayskips}
\appto{\small}{\zerodisplayskips}
\appto{\footnotesize}{\zerodisplayskips}

\fancypagestyle{firststyle}
{
\fancyhead{}

\fancyfoot[C]{\tiny © 2023 IEEE. Personal use of this material is permitted. Permission from IEEE must be obtained for all other uses, in any current or future media, including reprinting/republishing this material for advertising or promotional purposes, creating new collective works, for resale or redistribution to servers or lists, or reuse of any copyrighted component of this work in other works.}
}

\title{Leveraging Large Language Models for Exploiting ASR Uncertainty}
\name{Pranay Dighe$^*$, Yi Su$^*$, Shangshang Zheng, Yunshu Liu, Vineet Garg, Xiaochuan Niu, Ahmed Tewfik}
\address{Apple}

\begin{document}

\thispagestyle{firststyle}

\topmargin=0mm
\maketitle
\def\thefootnote{*}\footnotetext{Equal Contribution}

\vspace{-2mm}
\begin{abstract}
\vspace{-2mm}
While large language models excel in a variety of natural language processing (NLP) tasks, to perform well on spoken language understanding (SLU) tasks, they must either rely on off-the-shelf automatic speech recognition (ASR) systems for transcription, or be equipped with an in-built speech modality. This work focuses on the former scenario, where LLM’s accuracy on SLU tasks is constrained by the accuracy of a fixed ASR system on the spoken input. Specifically, we tackle speech-intent classification task, where a high word-error-rate can limit the LLM’s ability to understand the spoken intent. Instead of chasing a high accuracy by designing complex or specialized architectures regardless of deployment costs, we seek to answer how far we can go without substantially changing the underlying ASR and LLM, which can potentially be shared by multiple unrelated tasks. To this end, we propose prompting the LLM with an $n$-best list of ASR hypotheses instead of only the error-prone 1-best hypothesis. We explore prompt-engineering to explain the concept of $n$-best lists to the LLM; followed by the finetuning of Low-Rank Adapters~\cite{hu2021lora} on the downstream tasks. Our approach using $n$-best lists proves to be effective on a device-directed speech detection task as well as on a keyword spotting task, where systems using $n$-best list prompts outperform those using 1-best ASR hypothesis; thus paving the way for an efficient method to exploit ASR uncertainty via LLMs for speech-based applications.

\end{abstract}


\vspace{-3mm}
\section{Introduction}
\label{sec:intro}
\vspace{-2mm}
Large language models have recently revolutionized the field of NLP by showing excellent performance on a diverse set of downstream text-processing tasks often with little to no finetuning required on the downstream tasks~\cite{brown2020neurips,ouyang2022training,chen2021evaluating, bang2023multitask,qin2023chatgpt}. When trained with a large number of parameters in the order of billions (and even trillions~\cite{openai2023gpt4}) and similarly large quantities of text data, these models demonstrate an emergent ability to do in-context learning~\cite{xie2021explanation, min2022rethinking} and reasoning via chain-of-thought prompting~\cite{wei2023chainofthought}, which renders LLMs more accurate than dedicated smaller models trained on task-specific data.

\begin{figure}
    \captionsetup{font=footnotesize}
    \centering
    \includegraphics[width=0.9\linewidth]{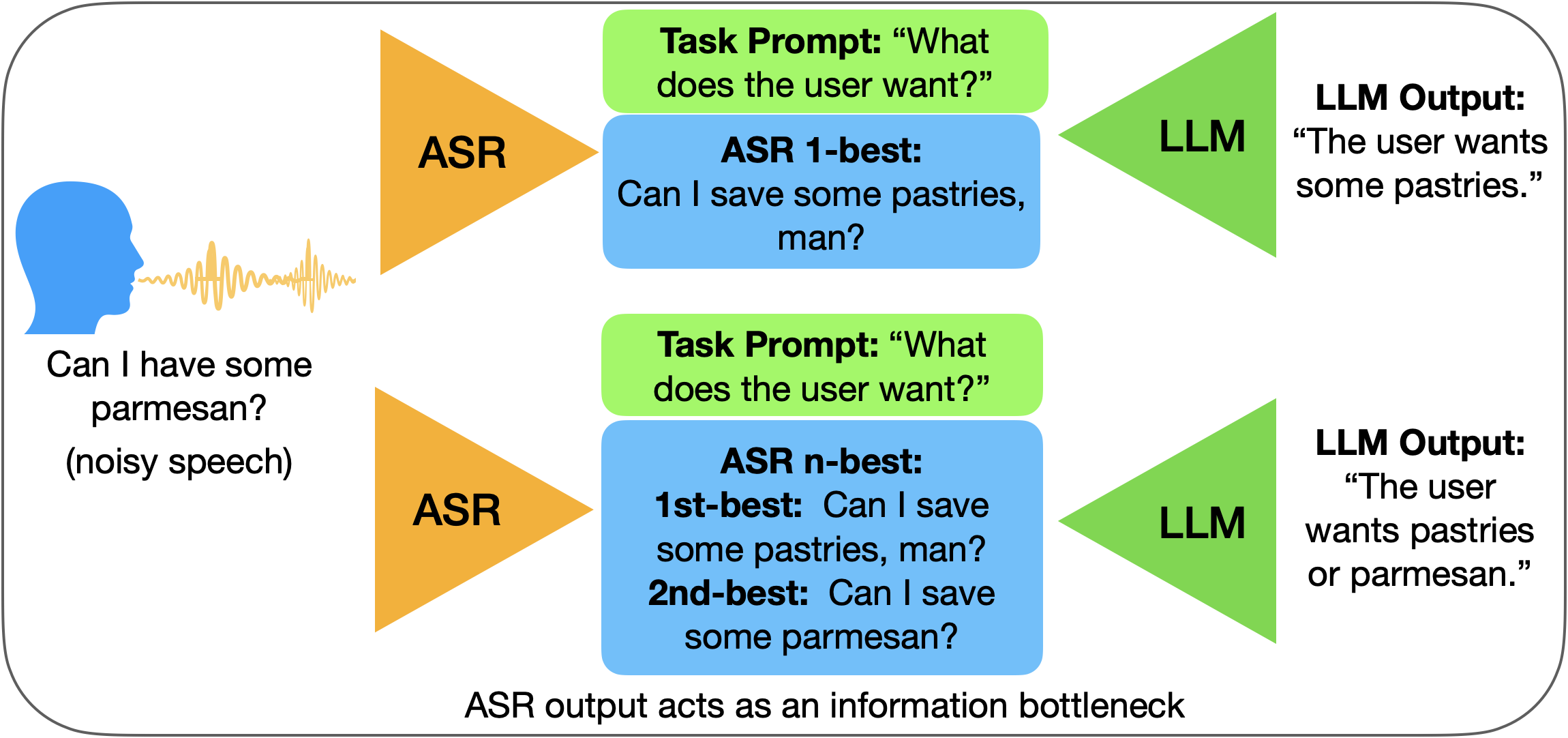}
    \caption{A toy NLP task which demonstrates that prompting the LLM with $n$-best ASR hypotheses allows it to exploit ASR uncertainty to better solve the downstream NLP task.}
    \label{fig:motivation}
\end{figure}

While this out-of-the-box generalizability and robustness of LLMs have instantly made them a popular tool for text-based applications~\cite{openai2023chatgpt,googlebard}, their usage for traditional speech-based applications is still an ongoing topic of research. One set of approaches relies on multi-modal LLMs, which ingest the audio modality by processing the underlying speech using an encoder network and feeding the LLM with speech embeddings~\cite{gong2023listen, fathullah2023prompting,zhang2023speechgpt,deshmukh2023pengi}. However, such architectures are mainly motivated towards the goal of having a single end-to-end model and the speech recognition capabilities of such multi-modal LLMs are limited~\cite{gong2023listen}. Another set of approaches interface LLMs with pretrained ASR models such that the ASR output is fed directly to the LLM as a prompt to tackle the downstream task~\cite{huang2023audiogpt,shen2023hugginggpt,he2023chatgpt}. The benefit of this modular approach is that one can choose any desirable in-domain ASR model and any LLM of appropriate size and configuration. In this work, we focus on such modular ASR+LLM architecture to approach the speech intent classification task and address the problem of incorrect ASR outputs affecting the ability of LLMs to determine the correct underlying intent.

An important consideration regarding LLM-based solutions, that has not received enough attention in the literature, is the high cost of deploying them, either server-side or on-device. We argue that to viably deploy LLMs into production, multiple tasks need to utilize one underlying LLM, effectively sharing the cost. Therefore, it is crucial to seek out ways of using LLMs in a non-intrusive, shareable way. This consideration has informed the scope of our exploration. For example, off-the-shelf frozen ASR systems and LLMs are preferred over complex specialized models; prompting, if effective, is preferred over finetuning; and finally, LoRA finetuning is favored over full finetuning of LLMs.

\begin{table*}[t!]
\small
\centering
\resizebox{\textwidth}{!}{%
\begin{tabular}{p{10.0cm}p{1.2cm}p{8cm}}
 \rowcolor{gray!50}
 \hline
\multicolumn{1}{c}{Prefix} & \multicolumn{1}{c}{Infix} &  \multicolumn{1}{c}{Suffix}\\\hline\hline
\textbf{1-best:} \textit{``Determine whether the following spoken utterance is directed towards a voice assistant or a human being.''} & \multicolumn{1}{c}{\multirow{2}{0.15\textwidth}[-2em]{\textit{``Typical spoken utterances directed towards the voice assistant are commands to fulfill a task or queries to get some information.''}}} & \textbf{binary-target:} \textit{``Answer only from the following categories [`1', `0'] where `1' indicates that the utterance is directed towards the voice assistant and `0' indicates that the utterance is directed towards a human being.''}
\\\cline{1-1}\cline{3-3}
\textbf{$n$-best:} \textit{``In this task, we provide an $n$-best list of ASR hypotheses for a spoken utterance. Each of the hypothesis is separated by a newline character. The cost of each hypothesis is at the end in the format `[cost]' where a low cost indicates that we are more confident about that ASR hypothesis. Determine whether the following spoken utterance is directed towards a voice assistant or a human being by taking into account all the $n$-best hypotheses.''} &  & \textbf{0-100 scale:} \textit{``Answer on a scale of 0 to 100 where a score of `100' indicates that the utterance is directed towards the voice assistant and `0' indicates that the utterance is directed towards a human being. Your answer should only contain an integer between 0 and 100.''}\\\hline
\end{tabular}
}
\caption{Task-prompts used for device-directed speech detection task.}
\label{table:ddsd_prompts}
\end{table*}

While ASR models aim to transcribe speech to text accurately, speech recognition on real-world speech is often inaccurate and word-error-rates (WER) are typically non-zero due to challenging conditions like noisy speech,  environmental noise, accents, acoustic and speaker variations~\cite{yu2016automatic}. For any downstream SLU task, the 1-best ASR hypothesis acts as an information bottleneck between the ASR and the LLM component, as shown in Figure \ref{fig:motivation}. In this work, we propose to widen this information bottleneck by exposing the LLM to an $n$-best list of ASR hypotheses. Our hypothesis is that using $n$-best lists instead of 1-best enables the LLM to benefit from the uncertainties in ASR prediction instead of being adversely affected by them. $n$-best lists also conform to our goal of making minimal changes to an LLM as they are a prompting-friendly format to convey ASR uncertainty to the LLM as discussed in Section \ref{sec:prompts}. Using descriptive prompts which explain the concept of $n$-best lists to an LLM, or by finetuning the LLM with $n$-best lists as the prompt, we demonstrate that an LLM can interface with a frozen ASR model in a more effective way as compared to using just the 1-best ASR output directly.
A relevant work~\cite{he2023chatgpt} utilizes ASR 1-best outputs as well as oracle transcripts as prompts to the LLM for speech-intent classification with encouraging results. Another work~\cite{ma2023generative} explores correcting ASR errors by prompting $n$-best lists to ChatGPT which results in improvements in ASR WER. Our work differs in multiple ways: (i) we explore prompting with $n$-best lists for the downstream intent classification and keyword spotting tasks, (ii) our $n$-best lists are augmented with ASR hypothesis costs as an additional source of information on uncertainty in ASR, and (iii) apart from direct prompting, we also explore finetuning of LoRA adapters using a training data of $n$-best list prompts. We use an in-house general-purpose English ASR system and Vicuna~\cite{vicuna2023}, an instruction-tuned LLaMA LLM~\cite{touvron2023llama} , in this paper, and we experiment on (i) a device-directed speech detection (DDSD) task for binary intent classification on whether an utterance is directed towards a voice assistant or not, and on (ii) a keyword spotting (KS) task on the Google Speech Commands (GSC)~\cite{warden2018speech} dataset, which we treat as a multi-class intent classification problem. While the labels for the binary classification DDSD task are typically 1 (\textit{directed}) and 0 (\textit{undirected}), we also explore the LLM's capability to output its decision on a scale of 0 and 100 so that the output score can be converted to a probability and used for generating smooth ROC curves.
We show improved accuracy using $n$-best lists on both GSC and DDSD tasks. 

The rest of the paper is structured as follows: Section \ref{sec:approach} explains our methodology. Section \ref{sec:experiments} provides experimental details and analysis. Section \ref{sec:conclusions} summarizes the conclusions.

\vspace{-4mm}
\section{Our Approach}
\label{sec:approach}
\vspace{-2mm}
In this section, we explain details of the SLU tasks and design of our inputs to the LLM.

\vspace{-4mm}
\subsection{Tasks and Datasets}
\label{sec:datasets}
\vspace{-2mm}

\noindent\textbf{Device-directed Speech Detection (DDSD)}: In this task, the goal is to identify whether a spoken utterance is directed towards a device (e.g. smartphone) or a human. We use an internal dataset which contains a \textit{train} and an \textit{eval} partition with a label of $1$ for device-directed speech and a label of $0$ for human-directed speech. The \textit{train} partition contains weakly-labeled data as explained in~\cite{garg2022ddsd} and comprises $\sim$107k utterances for each class. The \textit{eval} partition contains human-graded data with 12,771 device-directed and 2,274 human-directed utterances. We use our ASR+LLM intent classification system to predict the binary target labels on the eval set. In another set of experiments, we generate probabilistic scores on the \textit{train} partition using a LatticeRNN model~\cite{Ladhak2016} in range of $[0, 1]$. The floating point probability scores are then converted to an integer range $[0, 100]$ by multiplying with $100$ and rounding to the nearest integer. During LoRA finetuning, we train the LLM to output either the binary labels for the input utterance or a score in the scale of 0 to 100. In the latter case, the score between 0 and 100 is divided by 100 to get a pseudo-probability value\footnote{Another way to obtain probabilistic DDSD scores is using the Softmax probability of the tokens `1'/`0' in the LLM output layer. However, we choose to explore limits of interacting with LLM via prompting and text-generation only.}.
\\
\noindent\textbf{Keyword Spotting:} The GSC dataset consists of 35 keywords where 10 keywords \textit{``Yes'', ``No'', ``Up'', ``Down'', ``Left'', ``Right'', ``On'', ``Off'', ``Stop'',} and \textit{``Go''} are considered in-domain commands and the remaining words are considered out-of-vocabulary (OOV). Given an audio of 1 second duration, the task is to identify which command keyword, if any, was spoken in the utterance. State-of-the-art approaches~\cite{Kim2021BroadcastedRL,berg2021keyword,vygon2021learning,seo2021wav2kws}  typically use a discriminative classifier trained to identify the keywords-of-interest and achieve $\sim$98.5\% accuracy on this task. In this paper, we use a general-purpose large-vocabulary ASR system in conjunction with an LLM to identify the command keywords. Our goal remains to modify existing ASR and LLM solutions minimally, and we seek to explore the possibilities of what an LLM can do with uncertain outputs from an ASR system which is not trained discriminatively on this limited vocabulary KS task. The LLM in this case is expected to rectify the ASR output and map it to the correct keyword. We use the \textit{test} partition ($\sim$11k utterances) of this dataset for evaluating our ASR+LLM approach and the \textit{train} partition ($\sim$85k utterances) for LoRA finetuning of the LLM. More details of this dataset are available in~\cite{warden2018speech}.

\begin{table}[h!]
\centering
\renewcommand{\arraystretch}{1}
\resizebox{0.9\columnwidth}{!}{%
\begin{tabular}{lcc}
 \rowcolor{gray!50}
 \hline
utterance-prompt & GSC & DDSD \\\hline\hline
1-best & \textit{``hive''} & \textit{``shuffle play U2''}\\\hline
$n$-best & \makecell{\textit{``hive [-47.8]}\\ \textit{five [-46.8]} \\ \textit{bye [-44.0]} \\ \textit{hive [-31.5]''} } & \makecell{\textit{``shuffle play U2 [-84.4]}\\\textit{shuffle play Kito [-83.1]}\\\textit{shuffle play Buku [-82.9]}\\\textit{shuffle play Kitu [-82.8]''}} \\\hline
Ground-truth & \textit{``five''} & \textit{``shuffle play U2''}\\\hline
\end{tabular}
}
\caption{Examples of 1-best vs. $n$-best lists utterance-prompts.}
\label{table:nbest_examples}
\end{table}

\vspace{-4mm}
\subsection{ASR Outputs as $n$-best Lists}
\label{sec:nbest}
\vspace{-2mm}

ASR systems typically output a \textit{lattice} graph comprising competing ASR hypotheses under a beam-search decoder~\cite{mohri2002weighted}. The least cost path in the lattice is the 1-best hypothesis. While the full lattice may contain more paths than the $n$-best list and it is a richer format for capturing ASR uncertainty, we found that expressing lattices as prompts for LLM is not trivial because they are a complex data-structure whose textual representation is very long, often exceeding the maximum sequence length of 2048 tokens allowed for our base LLM model. Therefore, we choose $n$-best lists as the medium to inform the LLM of uncertainty in the ASR decoding process. An $n$-best list of competing ASR hypotheses can be obtained by picking the $n$ least cost paths in the lattice. We use a prompting-friendly format for the $n$-best lists where the hypotheses are separated with newline characters and each hypothesis is appended with a \textit{hypothesis-cost} at the end in the format \textit{[cost]}. This cost is the sum of the acoustic-model and language-model costs on the arcs along the lattice-path corresponding to the hypothesis~\cite{povey2012lattice} and a low cost indicates a high posterior probability assigned by the ASR system for that hypothesis given the speech input. We obtain such $n$-best lists for both DDSD and GSC datasets. The 1-best ASR hypothesis and the $n$-best lists are utterance-specific and we call them the \textit{utterance-prompt}. Table \ref{table:nbest_examples} shows examples of such utterance-prompts.

\begin{table*}[t!]
\centering
\setlength{\tabcolsep}{3pt}
\resizebox{\textwidth}{!}{
\renewcommand{\arraystretch}{1}
\begin{tabular}{c|cc|cc|cc|cc||cc|cc|cc|cc|cc}
\hline
& \multicolumn{8}{c||}{Binary Target} &  \multicolumn{10}{c}{100-Scale Task}\\\cline{2-19}
n & \multicolumn{2}{c|}{Base Model} & \multicolumn{2}{c|}{Finetuned} & \multicolumn{2}{c|}{*FT no-TP} & \multicolumn{2}{c||}{*FT no-HC} &  \multicolumn{2}{c|}{Base Model} & \multicolumn{2}{c|}{Finetuned} & \multicolumn{2}{c|}{*FT no-TP} & \multicolumn{2}{c|}{*FT GibTP} & \multicolumn{2}{c}{*FT no-HC} \\\cline{2-19}
& TPR $\uparrow$ & FPR $\downarrow$ & TPR $\uparrow$ & FPR $\downarrow$ & TPR $\uparrow$ & FPR $\downarrow$ & TPR $\uparrow$ & FPR $\downarrow$ & FPR95 $\downarrow$ & EER $\downarrow$ & FPR95 $\downarrow$ & EER $\downarrow$  & FPR95 $\downarrow$ & EER $\downarrow$ & FPR95 $\downarrow$ & EER $\downarrow$ & FPR95 $\downarrow$ & EER $\downarrow$ \\\hline\hline
1 & 91.5 & 30.0 & 90.5 & 8.3 & 90.3 & 8.6 & 90.5 & 8.3 & 84.7 & 32.0 & 55.8 & 10.7 & 54.0 & 9.9 & 54.5 & 10.3 & 53.6 & 10.7 \\
2 & 85.1 & 32.7 & 91.3 & 5.2 &  91.3 & 5.8 & 91.2 & 5.6 & 90.3 & 55.2 & 27.7 & 8.2 & 25.9 & 8.2 & 27.3 & 8.2 & 38.5 & 8.9 \\
4 & 87.2 & 46.4 & 91.8 & 5.2 &  91.8 & 5.2 & 91.5 & 5.2 & 85.8 & 46.0 & 13.0 & 7.6 & 12.0 & 7.5 & 12.2 & 7.5 & 21.1 & 7.7\\
8 & 85.9 & 43.0 & 91.8 & 4.7 & 91.9 & 5.3 & 91.4 & 4.9 & 81.9 & 36.3 & 11.0 & 7.5 & 11.4 & 7.2 & 11.0 & 7.3 & 12.3 & 7.3\\\
16 & 85.0 & 42.0 & 92.0 & 4.9 & 91.8 & 5.3 & 91.7 & 5.1 & 84.7 & 36.7 & 10.1 & 7.4 & 10.9 & 7.0  & 10.6 & 7.2 & 10.5 & 7.0\\\hline
\end{tabular}
}
\caption{Comparing accuracy of various ASR+LLM model setups on DDSD task with binary targets as outputs vs. output on a scale of 0-100 for different values of $n$-best list size. All numbers are in percentage. FPR95 refers to FPR of 100-Scale system at TPR=95\%. * refers to ablation studies on finetuning (FT) without task-prompt (no-TP), without hypothesis-cost (no-HC) or with gibberish task-prompts (GibTP) as compared to the ``Finetuned'' system.}
\label{table:ddsd_results}
\end{table*}

\vspace{-4mm}
\subsection{Designing Prompts for Processing $n$-best Lists}
\label{sec:prompts}
\vspace{-2mm}

We use the prompting-without-finetuning approach only for the DDSD task. Prompts that describe the DDSD task to the LLM are presented in Table \ref{table:ddsd_prompts}. We call these fixed prompts the \textit{task-prompt}. Each task-prompt is composed of a prefix, an infix, and a suffix to cover variations such as 1-best vs. $n$-best input and binary scoring vs. scoring on a scale of 0-100. The task-prompt is further concatenated with the DDSD utterance-prompts illustrated in Table \ref{table:nbest_examples}. For the KS task, we only rely on the finetuning approach as a pure prompting-without-finetuning approach could not beat our baseline KS system described in Section \ref{sec:baselines}.

\vspace{-3mm}
\section{Experiments}
\label{sec:experiments}
\vspace{-2mm}
In this section, we provide details of models, training, experiments and subsequent analysis.

\vspace{-2mm}
\vspace{-2mm}
\subsection{ASR System}
\vspace{-2mm}
Our ASR model is based on an E2E-ASR architecture with same hyperparameter setting as presented in~\cite{wu2021u2++}. The same model is used for both DDSD and GSC task. The model comprises a Conformer~\cite{gulati2020conformer} encoder with a CTC and an attention-based decoder. Beam-search decodings from the attention-based decoder are rescored using an external finite state transducer (FST) based language model. We use the ASR lattices obtained from the FST-LM decoding to generate the 1-best and the $n$-best ASR hypotheses. This model is trained for 100 epochs on $\sim$18k hours of speech data. For experiments in this work, the ASR model is considered frozen and is used only for generating the ASR outputs.  

\vspace{-4mm}
\subsection{LLM Base Model and Finetuning}
\vspace{-2mm}
We use a pretrained instruction-tuned LLM, Vicuna-7B-v1.3~\cite{vicuna2023}, as the base model for our experiments. Inference on this model is done using 4 NVIDIA A100 GPUs. For finetuning, we train parameters of LoRA adapters~\cite{hu2021lora} for 3 epochs using 8 GPUs with a learning rate of 2e-5 which we warmup to over 3\% of the learning steps. The LoRA adapters (rank=8) have 4.1M parameters which is 0.06\% of the 7B parameters of the base model. We use FastChat toolkit~\cite{zheng2023judging} for inference and finetuning with DeepSpeed GPU optimization~\cite{rasley2020deepspeed}. All inferences are deterministic with a temperature of 0. Finetuning is done using \textit{train} partitions of GSC and DDSD datasets.

\vspace{-4mm}
\subsection{Baseline Systems}
\label{sec:baselines}
\vspace{-2mm}
For the DDSD task, we use the LatticeRNN model~\cite{jeon2019latrnn} as our baseline. LatticeRNN is a strong baseline system as it processes the whole ASR decoding lattice whereas $n$-best list used in proposed ASR+LLM approach is a condensed prompting-friendly feature derived from the lattice. We also use the LatticeRNN to \textit{``teach''} the ASR+LLM system to output its decision on a scale of 0 to 100 and LatticeRNN could also be seen as a \textit{teacher} model under the \textit{student-teacher} knowledge-transfer paradigm. For the KS task, we devise a trivial baseline which claims that a keyword was detected if the ASR outputs that keyword as the 1-best hypothesis. If the ASR 1-best hypothesis does not match one of the 10 command keywords, we classify the utterance as OOV. We expect our ASR+LLM approach to improve upon this trivial baseline by rectifying the ASR predictions using the information in the $n$-best list prompts.

\vspace{-3.5mm}
\section{Results and Analysis}
\vspace{-2mm}
In this section, we provide experimental results and subsequent analysis using ablation studies.

\vspace{-5mm}
\subsection{Output Generations from Base Model vs. Finetuned Model}
\label{sec:finetuning}
\vspace{-2mm}
We aim to exploit the emergent ability of the LLM to \textit{understand} the task based on the provided prompt and output an appropriate answer. Towards this end, we prompt the ASR+LLM system directly with the task-prompt plus the utterance prompt without any finetuning. We observed that LLM can indeed perform the intent detection task out-of-the-box, but sometimes it may output a descriptive answer in natural language as opposed to a binary 0/1 output as desired for the DDSD task. While this behaviour demonstrates the emergent ability of the base LLM model, we consider it undesirable in this work because descriptive answers are not suitable for scoring the model accuracy easily. For example, we obtain the following outputs from the Vicuna-7B-v1.3 model with and without finetuning:

\begin{mytcolorbox}
\textbf{DDSD Prompt:} \textit{Task-prompt ($n$-best Prefix + Infix + binary-target Suffix) + ``score of the Hawaii game [-144.2] \textbackslash n what's the score of the Hawaii game [-135.1]''}\\
\textbf{LLM without Finetuning:} \textit{``Based on the provided $n$-best list, the most likely hypothesis for the spoken utterance is: `score of the Hawaii game [-144.2]'. This hypothesis is the first in the list, indicating that it is the most likely interpretation of the spoken utterance. The cost associated with this hypothesis is `-144.2', which suggests that the ASR system is not very confident about this hypothesis. Given the content of the utterance, it is more likely to be directed towards a voice assistant rather than a human being. The utterance contains a question about the Hawaii game, which is a sports-related topic that is commonly addressed to voice assistants. Therefore, the answer is `1'.''}\\
\textbf{LLM with Finetuning:} \textit{``1''}
\end{mytcolorbox}


As a convention, if the LLM outputs a descriptive answer in the DDSD task, we consider the underlying utterance to be classified as \textit{device-directed}. We found the LLM to output descriptive answers only for $\sim$6\% of the utterances when it was prompted without finetuning. With finetuning, the LLM outputs only the desired values and no descriptive answers.

\begin{table*}[tp]
\renewcommand{\arraystretch}{1}
\setlength{\tabcolsep}{6pt}
\resizebox{\textwidth}{!}{
\centering
\begin{tabular}{c|cccccccccccc}
\hline
\textit{System} & \textbf{yes} & \textbf{no} & \textbf{up} & \textbf{down} & \textbf{left} & \textbf{right} & \textbf{on} & \textbf{off} & \textbf{stop} & \textbf{go} & \textbf{OOV} & \textbf{Total Acc.} \\\hline\hline
Baseline & 98.6/98.1 & 83.4/98.3 & 99.6/55.1 & 100.0/87.2 & 100.0/88.3 & 100.0/87.1 & 100.0/78.5 & 99.5/90.3 & 99.0/98.8 & 99.7/81.3 & 93.0/99.3 & 94.5\\\hline
FT n=1 & 98.3/98.6 & 94.4/95.6 & 97.8/84.5 & 99.5/89.4 & 99.8/95.2 & 99.7/97.5 & 98.2/82.3 & 96.4/93.3 & 99.0/99.0 & 97.9/91.3 & 96.4/99.5 & 97.0\\
FT n=8 & 99.3/99.0 & 94.1/98.0 & 97.8/85.7 & 99.5/92.1 & 99.0/95.2 & 99.7/98.2 & 98.5/83.8 & 97.7/93.5 & 99.3/98.8 & 99.0/94.8 & 96.9/99.5 & 97.5 \\
\hline
\end{tabular}
}
\caption{Precision/Recall of various ASR+LLM models on keyword spotting task on GSC dataset for each keyword. All values are in percentage. Last column shows total accuracy.}
\label{tab:gsc_results}
\end{table*}

\vspace{-4mm}
\subsection{Intent Classification on DDSD Dataset}
\vspace{-2mm}
Table \ref{table:ddsd_results} shows the results of DDSD experiments. When we prompt or finetune the LLM to output a binary decision (`0' or `1') for DDSD utterance prompts, the output of ASR+LLM system provides a fixed True Positive Rate (TPR) and False Positive Rate (FPR), and the system is not tunable. Directly prompting the base model with task-prompts in Table \ref{table:ddsd_prompts} depicts the LLM's internal knowledge and emergent ability to tackle the DDSD task. We found that the base model \textit{understands} the task best when it is prompted with 1-best lists. It correctly rejects 70.0\% of the false positives at a high TPR of 91.5\%. As the size of the $n$-best list is increased, the model does not achieve as high TPR and as low FPR as the $n$=1 case, thus, suggesting that the base model is not acquainted with the concept of $n$-best lists from its original training and it is not able to exploit the ASR uncertainty as we desire.
When we LoRA finetune the base model to output binary targets, the accuracy on DDSD task improves considerably with TPR being consistently greater than 90\% for all values of $n$ in the $n$-best list. The system with 1-best ASR hypothesis has an FPR of only 8.6\% whereas with an 8-best list, it reduces to 4.7\% depicting that the ASR+LLM can correctly reject 95.3\% human-directed utterances while correctly predicting 91.8\% of the device-directed utterances. While we observe diminishing gains as we increase the size of $n$-best list, these observations provide strong evidence that LLMs can effectively use $n$-best lists for DDSD task.

\begin{figure}[h]
    \centering
    \includegraphics[width=0.75\linewidth, height=0.6\linewidth]{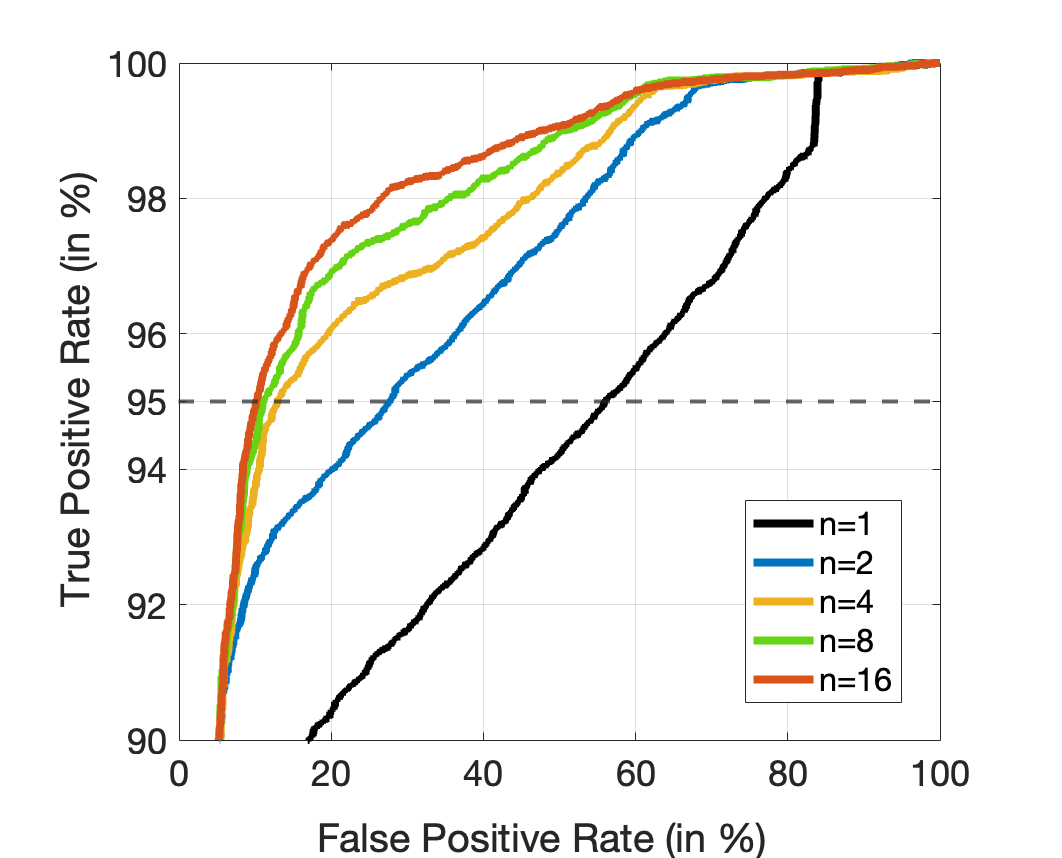}
    \caption{ROC curves for LLMs finetuned to output score on a scale of 0-100.}
    \label{fig:roc_ddsd}
\end{figure}

Next, we prompt and finetune the LLM to output scores on a scale of 0-100. In this \textit{100-Scale} task, the LLM outputs can actually be converted to a pseudo-probability by dividing the text output by 100. Therefore, this system is tunable and we evaluate it using Equal Error Rates (EER) and an arbitrarily chosen operating point of high TPR=95\%. Without finetuning, we consider this a very challenging task for the base model as the LLM is essentially expected to deduce a probability of an utterance being device-directed with a precision of 2-decimal places. As expected, the base model performs very poorly on this task for all values of $n$. When we LoRA finetune the base model for this task, the LLM accuracy improves substantially as shown in Table \ref{table:ddsd_results}. For 1-best ASR hypothesis, the EER improves from 32.0\% to 10.7\% and it improves to an EER of 7.4\% with $n$=16. As these systems are tunable, we evaluate them at TPR=95\%, a high TPR which is unachievable using the system which outputs binary targets only. The FPR decreases from 53.6\% for $n$=1 to 10.5\% for $n$=16, a relative reduction of 80\%. Figure \ref{fig:roc_ddsd} shows the ROC curves for the finetuned LLMs on the 100-scale task. $n$=16 system is the best performing system; and the area-under-the-curve consistently increases as $n$ is increased. In comparison to our ASR+LLM approach, the LatticeRNN baseline has an EER of 6.4\% and FPR=8.2\% at TPR=95\% which represents an upper bound of accuracy as LatticeRNN processes the full ASR lattice.

\textbf{Ablation Study on Prompts:} We ablated the utterance-prompts and task-prompts to determine their importance in solving the DDSD task. When hypothesis-costs are removed from the $n$-best lists, we found that the prompting-based approach on the base LLM model completely breaks down, and we couldn't engineer any effective task-prompt that achieves a reasonable accuracy on the DDSD task. However, when the LLM is LoRA finetuned, we found that neither the removal of the hypothesis-cost nor that of the task-prompts has significant impact on the model accuracies as shown in Table \ref{table:ddsd_results}. Furthermore, we also LoRA finetuned the LLM using a gibberish task-prompt composed of non-English words and obtained fairly similar accuracy as the original prompt. The ablation studies show that only the utterance-prompts (without hypothesis costs) are required for LoRA finetuning.

\vspace{-2mm}
\subsection{Keyword Spotting on GSC Dataset}
\vspace{-1mm}
Table \ref{tab:gsc_results} shows the results for the KS task on the GSC dataset for the trivial ASR-based baseline and the ASR+LLM approach where the LLM is LoRA finetuned with prompting using only the best ASR hypothesis ($n$=1) versus prompting with $n$-best list with $n$=8. The ASR baseline has an overall accuracy of 94.5\% with low-recall for certain commands e.g. ``on'' and ``up''. When we feed the 1-best ASR hypothesis to the LLM and finetune for the KS task, the overall accuracy improves to 97.0\%. Examples of some corrections made by the LLM are \textit{``app''$\rightarrow$``up'', ``Lyft''$\rightarrow$``left''}, and \textit{``call''$\rightarrow$``go''}; and such corrections are learned by the LLM from the training dataset during finetuning. The LLM learns the common mistakes made by the ASR in the 1-best hypothesis and rectifies them. When we LoRA finetune the LLM with $n$-best list of size 8, the overall accuracy increases to 97.5\%. This improvement stems for the fact that $n$-best lists often contain the ground-truth keyword as one of the alternate hypothesis e.g. \textit{``out [-42.7] \textbackslash n app [-42.6]\textbackslash n OK [-41.3] \textbackslash n home [-40.9] \textbackslash n oh [-40.1] \textbackslash n no [-39.7] \textbackslash n hello [-38.0] \textbackslash n up [-37.8]''} is correctly classified as the keyword \textit{``up''} for an utterance with \textit{``up''} as the ground-truth. In this example, while the correct command \textit{``up''} is 7th in the $n$-best list, the LLM is able to extract it based on the various alternate hypotheses.

\vspace{-5mm}
\section{Conclusions}
\vspace{-4mm}
\label{sec:conclusions}
In this work, we explore prompting large language models with $n$-best ASR hypotheses to tackle downstream tasks like spoken intent classification and keyword spotting. We hypothesize that LLMs can better exploit the uncertainty in ASR by processing $n$-best lists instead of the 1-best ASR outputs which are highly prone to errors. Our experiments show moderate success on the intent classification task when the base LLM model is directly prompted with descriptive task-prompts, with no clean advantage of using $n$-best lists. However, we demonstrate significant improvements on both the tasks using $n$-best lists with LoRA finetuning and confirm our initial hypothesis that LLMs can indeed leverage rich ASR information. Specifically, on the intent classification task, we design a tunable system which outputs its decision on a scale of 0 to 100 enabling us to tune for any desired operating point. In future, the proposed approach can be generalized to more complex SLU tasks like domain prediction and intent classification for multi-turn dialogues.

\vspace{-4mm}
{\footnotesize 
\section{Acknowledgements}
\vspace{-4mm}
We thank John Bridle, Shruti Palaskar, and Barry Theobald for their feedback on the paper.}

{\footnotesize\bibliography{refs}}

\begin{thebibliography}{10}

\bibitem{hu2021lora}
E.~J. Hu, Y.~Shen, P.~Wallis, Z.~Allen-Zhu, Y.~Li, S.~Wang, L.~Wang, and
  W.~Chen, ``Lo{RA}: Low-rank adaptation of large language models,'' in {\em
  International Conference on Learning Representations}, 2022.

\bibitem{brown2020neurips}
T.~Brown, B.~Mann, N.~Ryder, M.~Subbiah, J.~D. Kaplan, P.~Dhariwal,
  A.~Neelakantan, P.~Shyam, G.~Sastry, A.~Askell, S.~Agarwal, A.~Herbert-Voss,
  G.~Krueger, T.~Henighan, R.~Child, A.~Ramesh, D.~Ziegler, J.~Wu, C.~Winter,
  C.~Hesse, M.~Chen, E.~Sigler, M.~Litwin, S.~Gray, B.~Chess, J.~Clark,
  C.~Berner, S.~McCandlish, A.~Radford, I.~Sutskever, and D.~Amodei, ``Language
  models are few-shot learners,'' in {\em Advances in Neural Information
  Processing Systems} (H.~Larochelle, M.~Ranzato, R.~Hadsell, M.~Balcan, and
  H.~Lin, eds.), vol.~33, pp.~1877--1901, Curran Associates, Inc., 2020.

\bibitem{ouyang2022training}
L.~Ouyang, J.~Wu, X.~Jiang, D.~Almeida, C.~Wainwright, P.~Mishkin, C.~Zhang,
  S.~Agarwal, K.~Slama, A.~Ray, {\em et~al.}, ``Training language models to
  follow instructions with human feedback,'' {\em Advances in Neural
  Information Processing Systems}, vol.~35, pp.~27730--27744, 2022.

\bibitem{chen2021evaluating}
M.~Chen, J.~Tworek, H.~Jun, Q.~Yuan, H.~P. d.~O. Pinto, J.~Kaplan, H.~Edwards,
  Y.~Burda, N.~Joseph, G.~Brockman, {\em et~al.}, ``Evaluating large language
  models trained on code,'' {\em arXiv preprint arXiv:2107.03374}, 2021.

\bibitem{bang2023multitask}
Y.~Bang, S.~Cahyawijaya, N.~Lee, W.~Dai, D.~Su, B.~Wilie, H.~Lovenia, Z.~Ji,
  T.~Yu, W.~Chung, {\em et~al.}, ``A multitask, multilingual, multimodal
  evaluation of {ChatGPT} on reasoning, hallucination, and interactivity,''
  {\em arXiv preprint arXiv:2302.04023}, 2023.

\bibitem{qin2023chatgpt}
C.~Qin, A.~Zhang, Z.~Zhang, J.~Chen, M.~Yasunaga, and D.~Yang, ``Is {C}hat{GPT}
  a general-purpose natural language processing task solver?,'' 2023,
  arXiv:2302.06476.

\bibitem{openai2023gpt4}
OpenAI, ``{GPT-4} technical report,'' 2023, arXiv:2303.08774.

\bibitem{xie2021explanation}
S.~M. Xie, A.~Raghunathan, P.~Liang, and T.~Ma, ``An explanation of in-context
  learning as implicit {B}ayesian inference,'' {\em arXiv preprint
  arXiv:2111.02080}, 2021.

\bibitem{min2022rethinking}
S.~Min, X.~Lyu, A.~Holtzman, M.~Artetxe, M.~Lewis, H.~Hajishirzi, and
  L.~Zettlemoyer, ``Rethinking the role of demonstrations: What makes
  in-context learning work?,'' 2022, arXiv:2202.12837.

\bibitem{wei2023chainofthought}
J.~Wei, X.~Wang, D.~Schuurmans, M.~Bosma, B.~Ichter, F.~Xia, E.~Chi, Q.~Le, and
  D.~Zhou, ``Chain-of-thought prompting elicits reasoning in large language
  models,'' 2023, arXiv:2201.11903.

\bibitem{openai2023chatgpt}
OpenAI, ``{ChatGPT}: Conversational {AI} powered by {GPT}-3.5,'' 2023.

\bibitem{googlebard}
Google{\ }AI, ``Google {Bard},'' 2023.

\bibitem{gong2023listen}
Y.~Gong, H.~Luo, A.~H. Liu, L.~Karlinsky, and J.~Glass, ``Listen, think, and
  understand,'' 2023, arXiv:2305.10790.

\bibitem{fathullah2023prompting}
Y.~Fathullah, C.~Wu, E.~Lakomkin, J.~Jia, Y.~Shangguan, K.~Li, J.~Guo,
  W.~Xiong, J.~Mahadeokar, O.~Kalinli, C.~Fuegen, and M.~Seltzer, ``Prompting
  large language models with speech recognition abilities,'' 2023,
  arXiv:2307.11795.

\bibitem{zhang2023speechgpt}
D.~Zhang, S.~Li, X.~Zhang, J.~Zhan, P.~Wang, Y.~Zhou, and X.~Qiu,
  ``Speech{GPT}: Empowering large language models with intrinsic cross-modal
  conversational abilities,'' 2023, arXiv:2305.11000.

\bibitem{deshmukh2023pengi}
S.~Deshmukh, B.~Elizalde, R.~Singh, and H.~Wang, ``Pengi: An audio language
  model for audio tasks,'' 2023, arXiv:2305.11834.

\bibitem{huang2023audiogpt}
R.~Huang, M.~Li, D.~Yang, J.~Shi, X.~Chang, Z.~Ye, Y.~Wu, Z.~Hong, J.~Huang,
  J.~Liu, Y.~Ren, Z.~Zhao, and S.~Watanabe, ``Audio{GPT}: Understanding and
  generating speech, music, sound, and talking head,'' 2023, arXiv:2304.12995.

\bibitem{shen2023hugginggpt}
Y.~Shen, K.~Song, X.~Tan, D.~Li, W.~Lu, and Y.~Zhuang, ``Hugging{GPT}:
  {S}olving {AI} {T}asks with {ChatGPT} and its {F}riends in {H}ugging
  {F}ace,'' 2023, arXiv:2303.17580.

\bibitem{he2023chatgpt}
M.~He and P.~N. Garner, ``Can {ChatGPT} detect intent? evaluating large
  language models for spoken language understanding,'' 2023, arXiv:2305.13512.

\bibitem{yu2016automatic}
D.~Yu and L.~Deng, {\em Automatic speech recognition}, vol.~1.
\newblock Springer, 2016.

\bibitem{ma2023generative}
R.~Ma, M.~Qian, P.~Manakul, M.~Gales, and K.~Knill, ``Can generative large
  language models perform {ASR} error correction?,'' 2023, arXiv:2307.04172.

\bibitem{vicuna2023}
W.-L. Chiang, Z.~Li, Z.~Lin, Y.~Sheng, Z.~Wu, H.~Zhang, L.~Zheng, S.~Zhuang,
  Y.~Zhuang, J.~E. Gonzalez, I.~Stoica, and E.~P. Xing, ``Vicuna: An
  open-source chatbot impressing {GPT}-4 with 90\%* {ChatGPT} quality,'' March
  2023.

\bibitem{touvron2023llama}
H.~Touvron, T.~Lavril, G.~Izacard, X.~Martinet, M.-A. Lachaux, T.~Lacroix,
  B.~Rozière, N.~Goyal, E.~Hambro, F.~Azhar, A.~Rodriguez, A.~Joulin,
  E.~Grave, and G.~Lample, ``{LLaMA}: Open and efficient foundation language
  models,'' 2023, arXiv:2302.13971.

\bibitem{warden2018speech}
P.~Warden, ``Speech commands: A dataset for limited-vocabulary speech
  recognition,'' 2018, arXiv:1804.03209.

\bibitem{garg2022ddsd}
V.~Garg, O.~Rudovic, P.~Dighe, A.~H. Abdelaziz, E.~Marchi, S.~Adya, C.~Dhir,
  and A.~Tewfik, ``Device-directed speech detection: Regularization via
  distillation for weakly-supervised models,'' in {\em Interspeech}, 2022.

\bibitem{Ladhak2016}
F.~Ladhak, A.~Gandhe, M.~Dreyer, L.~Mathias, A.~Rastrow, and B.~Hoffmeister,
  ``Lattice {RNN}: Recurrent neural networks over lattices,'' in {\em
  Interspeech 2016}, 2016.

\bibitem{Kim2021BroadcastedRL}
B.~Kim, S.~Chang, J.~Lee, and D.~Sung, ``Broadcasted residual learning for
  efficient keyword spotting,'' in {\em Interspeech}, 2021.

\bibitem{berg2021keyword}
A.~Berg, M.~O'Connor, and M.~T. Cruz, ``Keyword transformer: A self-attention
  model for keyword spotting,'' in {\em Interspeech 2021}, pp.~4249--4253,
  ISCA, 2021.

\bibitem{vygon2021learning}
R.~Vygon and N.~Mikhaylovskiy, ``Learning efficient representations for keyword
  spotting with triplet loss,'' in {\em Speech and Computer: 23rd International
  Conference, SPECOM 2021, St. Petersburg, Russia, September 27--30, 2021,
  Proceedings 23}, pp.~773--785, Springer, 2021.

\bibitem{seo2021wav2kws}
D.~Seo, H.-S. Oh, and Y.~Jung, ``Wav2kws: Transfer learning from speech
  representations for keyword spotting,'' {\em IEEE Access}, vol.~9,
  pp.~80682--80691, 2021.

\bibitem{mohri2002weighted}
M.~Mohri, F.~Pereira, and M.~Riley, ``Weighted finite-state transducers in
  speech recognition,'' {\em Computer Speech \& Language}, vol.~16, no.~1,
  pp.~69--88, 2002.

\bibitem{povey2012lattice}
D.~Povey, M.~Hannemann, G.~Boulianne, L.~Burget, A.~Ghoshal, M.~Janda,
  M.~Karafiát, S.~Kombrink, P.~Motlíček, Y.~Qian, K.~Riedhammer, K.~Veselý,
  and N.~T. Vu, ``Generating exact lattices in the {WFST} framework,'' in {\em
  2012 IEEE International Conference on Acoustics, Speech and Signal Processing
  (ICASSP)}, pp.~4213--4216, 2012.

\bibitem{wu2021u2++}
D.~Wu, B.~Zhang, C.~Yang, Z.~Peng, W.~Xia, X.~Chen, and X.~Lei, ``U2++: Unified
  two-pass bidirectional end-to-end model for speech recognition,'' {\em arXiv
  preprint arXiv:2106.05642}, 2021.

\bibitem{gulati2020conformer}
A.~Gulati, J.~Qin, C.-C. Chiu, N.~Parmar, Y.~Zhang, J.~Yu, W.~Han, S.~Wang,
  Z.~Zhang, Y.~Wu, and R.~Pang, ``Conformer: Convolution-augmented transformer
  for speech recognition,'' 2020, arXiv:2005.08100.

\bibitem{zheng2023judging}
L.~Zheng, W.-L. Chiang, Y.~Sheng, S.~Zhuang, Z.~Wu, Y.~Zhuang, Z.~Lin, Z.~Li,
  D.~Li, E.~P. Xing, H.~Zhang, J.~E. Gonzalez, and I.~Stoica, ``Judging
  {LLM}-as-a-judge with {MT}-bench and chatbot arena,'' 2023, arXiv:2306.05685.

\bibitem{rasley2020deepspeed}
J.~Rasley, S.~Rajbhandari, O.~Ruwase, and Y.~He, ``Deep{S}peed: System
  optimizations enable training deep learning models with over 100 billion
  parameters,'' in {\em Proceedings of the 26th ACM SIGKDD International
  Conference on Knowledge Discovery \& Data Mining}, KDD '20, (New York, NY,
  USA), p.~3505–3506, Association for Computing Machinery, 2020.

\bibitem{jeon2019latrnn}
W.~{Jeon}, L.~{Liu}, and H.~{Mason}, ``Voice trigger detection from {LVCSR}
  hypothesis lattices using bidirectional lattice recurrent neural networks,''
  in {\em ICASSP 2019 - 2019 IEEE International Conference on Acoustics, Speech
  and Signal Processing (ICASSP)}, pp.~6356--6360, May 2019.

\end{thebibliography}
\bibliographystyle{hieeetr}

\end{document}